\documentclass{article}

 \usepackage[preprint]{neurips_2019}

\usepackage{color}
\usepackage[utf8]{inputenc} 
\usepackage[T1]{fontenc}    
\usepackage{hyperref}       
\usepackage{url}            
\usepackage{booktabs}       
\usepackage{amsfonts}       
\usepackage{nicefrac}       
\usepackage{microtype}      
\usepackage{algorithm}
\usepackage{amsmath}
\usepackage[pdftex]{graphicx}

\newcommand{\mb}[1]{\mathbb{#1}}

\title{SelectNet: Learning to Sample from the Wild for Imbalanced Data
  Training}

\author{%
   Yunru~Liu \\
  Department of Mathematics \\
  National University of Singapore \\
  10 Lower Kent Ridge Road \\
  Singapore 119076 \\
  \texttt{e0189564@u.nus.edu} \\
  \And
  Tingran~Gao\\
  Committee on Computational and Applied Mathematics \\ 
  Department of Statistics \\
  University of Chicago \\
  5747 S Ellis Avenue Jones 316 \\
  Chicago IL 60637-1441 \\
  \texttt{tingrangao@galton.uchicago.edu} \\
  \And
  Haizhao~Yang \\
  Department of Mathematics and Institute of Data Science \\
  National University of Singapore \\
  10 Lower Kent Ridge Road \\
  Singapore 119076 \\
  \texttt{haizhao@nus.edu.sg} \\
}

\begin{document}

\maketitle

\begin{abstract}
Supervised learning from training data with imbalanced class sizes, a commonly encountered scenario in real applications such as anomaly/fraud detection, has long been considered a significant challenge in machine learning. Motivated by recent progress in curriculum and self-paced learning, we propose to adopt a semi-supervised learning paradigm by training a deep neural network, referred to as SelectNet, to selectively add unlabelled data together with their predicted labels to the training dataset. Unlike existing techniques designed to tackle the difficulty in dealing with class imbalanced training data such as resampling, cost-sensitive learning, and margin-based learning, SelectNet provides an end-to-end approach for learning from important unlabelled data ``in the wild'' that most likely belong to the under-sampled classes in the training data, thus gradually mitigates the imbalance in the data used for training the classifier. We demonstrate the efficacy of SelectNet through extensive numerical experiments on standard datasets in computer vision.
\end{abstract}



\section{Introduction}
\label{sec:introduction}

The success of supervised learning algorithms largely hinges upon high-quality training data. Due to resource constraints and the nature of the specific applications, it can often be difficult to train a classifier on a training data set with balanced numbers of samples within each class, especially in scenarios such as anomaly detection \citep{HA2004} and rare event discovery \citep{HGX2013}. in some instances, the difference between sample sizes within classes can differ by several orders of magnitude, easily causing serious inductive bias that leads to poor prediction performance for minor classes \citep{DGZ2019,buda2018systematic}. Unfortunately, often times in these applications it is much more important to successfully predict the minor class samples, such as disease discovery and fraud detection \citep{hospedales2013finding}.

Existing techniques for tackling the data imbalance problem can be roughly divided into two categories. One is to adopt balanced strategies for the class imbalanced training data, such as bootstrapping the minor classes or downsampling the major classes, or a combination of both, usually following an ensemble learning paradigm \citep{chawla2002smote,LWZ2009,MS2011}; the other is to adjust the learning objective, such as introducing different weights for samples from the major or minor classes respectively or employing a boosting strategy adapted to the heterogeneous sampling density \citep{ZLA2003,he2008learning}. The two categories of methods are not mutually exclusive --- in fact, it is often beneficial to combine the benefits of each type of methods to achieve even better results in practice. It is worth pointing out that, however, all these techniques are crafted towards fully exploiting the structure of the skewed training data, which suffers from the deficiency in the minor training data classes.

Inspired by the emerging trend of semi-supervised learning in the past decades, in this paper we propose to borrow powers from the unlabelled data ``in the wild,'' which are relatively easy to obtain (e.g. from modern search engines or web scrapers) but may be difficult or expensive to label (due to lack of time or human power). We hope to enlarge the training data set with more data instances of the minor classes, which balances out the skewness in the original training data, at the slight expense of incorporating few unlabelled data that are mistakenly treated as belonging to a minor class; we introduce an iterative learning strategy that is more reluctant at accepting a misclassified unlabelled data at beginning, but eventually gains more confidence and leverages the full power of unlabelled data to mitigate the imbalance issue in the original training data set. The gradual adjustment of the ``attitude'' towards unlabelled data is motivated by recent work in curriculum learning \citep{WCA2018,jiang2017mentornet} and self-paced learning \citep{KPK2010,JMZSH2015}; we implement this mechanism in an end-to-end fashion by means of a deep neural network, dubbed \emph{SelectNet}. We demonstrate using extensive numerical experiments that this architecture is capable of recognizing important samples from the unlabeled data that most effectively reduces the class imbalance issue in the original training data, in such a way that the minor class prediction accuracy gets improved at no expense of sacrificing the major class prediction accuracy.

In summary, the key contributions of this paper are as follows:

\begin{itemize}
\item Unlike existing techniques that strive to find an appropriate way to deal with the class imbalanced issues in the original data set, we propose the novel paradigm of leveraging the unlabelled data in a semi-supervised fashion;
\item We design an end-to-end deep neural network architecture, the SelectNet, which automatically learns to pick important data samples from the pool of unlabelled data and use them for improving the classifier;
\item The SelectNet can be realized as an additional regularization term for a deep neural network based classifier, which can be trained along with the main classification DNN in the same computational workflow;
\item Extensive numerical experiments are conducted to compare the performance of SelectNet against competing methods over standard computer vision benchmarks, and the results speak of the superior power of SelectNet in the face of severe class imbalance in the training data.
\end{itemize}



\section{Related Work}
\label{sec:prel-relat-work}

\subsection{Learning with Imbalanced Data}

\paragraph{Dataset Resampling} Two naive but effective way of resampling techniques are \emph{oversampling}, which repeatedly sample data from the minor class until reaching the desired amount,  and \emph{downsampling}, which sample the same amount of data from major class to match with minor class \citep{he2008learning,chawla2002smote,oquab2014learning}. When using traditional machine learning methods like linear classifiers, oversampling can cause serious overfitting \citep{chawla2002smote}. In the setting of deep neural networks, oversampling shows better compatibility while missing information in the downsampling strategy shows a critical disadvantage \citep{buda2018systematic}.

\paragraph{Cost-sensitive learning} Cost-sensitive learning strategies aims that adjusting the weights in the objective loss function for training samples from different classes. Popular weight adjusting strategies include assigning weights according to inverse class frequency \citep{huang2016learning,wang2017learning}, or with respect to the ``hardness'' of the training samples, e.g., those samples that are wrongly classified by the classifier being trained \citep{lin2017focal,dong2017class}. In a sense, resampling from origin class imbalanced training data set plays a similar role as assigning higher weights for the wrongly predicted samples.



\subsection{Semi-supervised Learning}
\paragraph{Self-paced Learning} In \citep{KPK2010}, the authors proposed \emph{self-paced learning}, an iterative approach to select ``easy'' training samples based on the current parameters of the neural network. The number of samples selected at each iteration is gradually annealed such that in the later learning stage well-trained model can learn more samples with better tolerance to noise. A related regime is co-training \citep{blum1998combining}, which alternately trains two or more classifiers, and passes ``confident training samples'' determined by one classifier to another classifier as training data, together with the ``confidently'' predicted label.

We give a slightly more detailed account of the paradigm of self-paced learning here, as this is an important motivation for our approach for tackling the imbalanced data issue. For a training data set $\mathcal{D} = \{(\textbf{x}_1, \textbf{y}_1),...,(\textbf{x}_n, \textbf{y}_n)\}$, self-paced learning uses a vector $\textbf{v}\in \left\{0,1\right\}^n$ to indicate whether or not each training sample should be included in the current training stage ($v_i = 1$ if the $i$th sample is included in the current iteration). The overall target function including $\textbf{v}$ at iteration t is
\begin{equation}\label{spl}
(\textbf{w}_{t+1}, \textbf{v}_{t+1})= \mathop{\arg\min}_{\textbf{w} \in \mb{R}^d, \textbf{v} \in \{0,1\}^n} \sum_{i=1}^n v_i\mb{L}(y_i, f(\textbf{x}_i, \textbf{w})) - \lambda \sum_{i=1}^n v_i
\end{equation}
where $\mb{L}(y_i, f(\textbf{x}_i, \textbf{w}))$ denote the loss function of a convolutional neural network (CNN) model and $\textbf{w}$ refer to the model weights. When this model is relaxed to $\textbf{v}\in [0,1]^n$, a straightforward derivation easily reveals a rule for the optimal value for each entry $v_i$ as
\begin{equation}
v_i = \left\{
  \begin{tabular}{cc}
  1 & $\mb{L}(y_i, f(\textbf{x}_i, \textbf{w})) < \lambda$,\\
  0 & \textrm{otherwise.}
  \end{tabular}
\right.
\end{equation}
In this formulation, the two sets of variables are model weights $\textbf{w}$ and indicator vector $\textbf{v}$, and a common training strategy is to alternatively fix one set of variables and optimize the other set.

\paragraph{MentorNet} The architecture of MentorNet was proposed in \citep{jiang2017mentornet} to further improve the loss thresholding strategy of self-paced learning. Instead of alternatively updating the ``curriculum,'' the authors suggest that we use a network to directly learn the weighing curriculum (the vector $\textbf{v}$) from the training data, which is trained on a subset of the training data that is cleaned up and labeled with either ``right'' or ``wrong,'' indicating whether the input classification label of one sample is the true label or manually disturbed one. The output vector of MentorNet is used as the weight for the loss of each training sample. MentorNet and the corresponding base net (StudentNet) are trained alternatively to provide better-labeled training data and improve the overall accuracy on noisy-labeled datasets. Inspired by its idea, we expanded our sample choosing vector into a network output, which will produce a better sign of confidence than the loss value used in self-paced training.

\paragraph{SPARC} \citet{zhou2018sparc} proposed the schema \emph{SPARC} for learning network representation from rare category data, with an additional unlabeled data set. The minor groups in networks are emphasized for generating good network representations. To capture the underlying distribution of rare category examples, the predictions of both labeled and unlabeled data are considered in weighing training samples, which jointly produce separable margins between minor groups and major groups. The idea of enhancing the margin also motivated us to reuse labeled data to secure boundary performance.

\section{Algorithm}
\label{sec:algorithm}

Motivated by the methodology of semi-supervised learning, the approach proposed here leverages both labeled and unlabeled data to mitigate the class imbalance issue. Intuitively, we would like to ``bootstrap'' the classifier by adding the unlabelled data predicted as the minor classes by the current classifier to the training set. This approach is similar to the ``pseudo-label'' approach used in practice \cite{Lee2013,WP2018}; the success of this procedure certainly relies on correctly identifying minor classes from the unlabeled data, and thus depends on the data distribution and the actual decision boundary. We propose to use SelectNet to learn which unlabelled data to add to the training set.

\subsection{Formulation}

We distinguish two sets of data, labeled and unlabelled, that are used for training. Denote $\mathcal{D}= (\textbf{x}_i,y_i)$ for the original labeled imbalanced dataset with $m$ classes, where $y_i$ are one-hot encodings of the class labels. Assume that $K$ out of the $m$ classes are deficient in class size and they will be referred to as the minor classes. We use $C_i$ to indicate the $i$th class, and $|C_i|$ for the number of training data in this class. The ratio $\frac{\max_i|C_i|}{\min_i|C_i|}$ measures the level of imbalance of the original training set. In addition, we assume an extra ``pool'' of unlabelled data $\mathcal{U}= (\textbf{x}_i)$ is available, from which we collect more training samples in the minor classes using the current classifier. In practice, these unlabelled data may be collected by keywords searching in google or crawled from the internet. The hypothesis on $\mathcal{U}$ is that they have a high potential of including data in minor classes.

For a classification task, denote its loss function as $\textbf{L}_c(y_i, f_c(\textbf{x}_i, \textbf{w}_c))$, where $f_c(\textbf{x}_i, \textbf{w}_c)$ is the main deep neural network with weights $\textbf{w}_c$ being trained for the classification task. In each iteration, the label of an unlabelled sample $\textbf{x}_i\in \mathcal{U}$ is inferred from the main classifier and is denoted as $\hat{y}_i= argmax(f_c(\textbf{x}_i, \textbf{w}_c))$. In the meanwhile, another neural network is trained to determine which of the unlabelled data will be added to the training sample for the next iteration. The second neural network is denoted as $f_s(\textbf{z}_i;\Theta)$ with parameters $\Theta$. $f_s(\textbf{z}_i;\Theta)$ takes as input the last layer feature of an unlabeled data output by $f_c$ along with its loss calculated by the classification model, i.e., $z_i=\left( f_c(\textbf{x}_i, \textbf{w}_c),\textbf{L} \left( f_c(\textbf{x}_i, \textbf{w}_c),\textbf{w}_c \right) \right)$. The entire model has the following form:
\begin{equation}\label{obj}
\begin{split}
\mathop{\min}_{\textbf{w} \in \mb{R}^d, f_s \in \{0,1\}^{n\times m}} \mb{F}(\textbf{w},\textbf{v}) &= \frac{1}{n_D} \sum_{i\in \mathcal{D}} \textbf{L}_c(y_i, f_c(\textbf{x}_i, \textbf{w}_c))\\
&+ \frac{1}{n_{add}} \sum_{i\in \mathcal{D}'\cup\mathcal{U}'} \left[f_s(\textbf{z}_i;\Theta)\textbf{L}_c(\hat{y}_i, f_c(\textbf{x}_i, \textbf{w}_c)) - \lambda f_s(\textbf{z}_i;\Theta)\right],
\end{split}
\end{equation}
where $\mathcal{D}'$ and $\mathcal{U}'$ include the data whose top-$1$ prediction result is a minor class. The training procedure alternatively updates the main classification network $f_c(\textbf{x}_i, \textbf{w}_c)$ and the selection network $f_s(\textbf{z}_i;\Theta)$. We remark that the ``selection'' part includes an additional regularization term with a penalty parameter $\lambda$, since $f_s$ will trivially attain zero value at all unlabelled data if $\lambda=0$. Moreover, only samples which are confidently predicted in a minor class are added to the training set so as to prevent the model from further aggravating data imbalance. Evaluation of the confidence differs from method to method. For example, in self-paced learning, loss value is directly used as a sign of confidence. The samples with loss value smaller than a hyperparameter $\lambda$ will be selected. We will introduce our indicator of confidence Section 3.3.


\subsection{Context Data}
\label{sec:cd}

\begin{figure}
  \centering
  \includegraphics[width=0.8\linewidth]{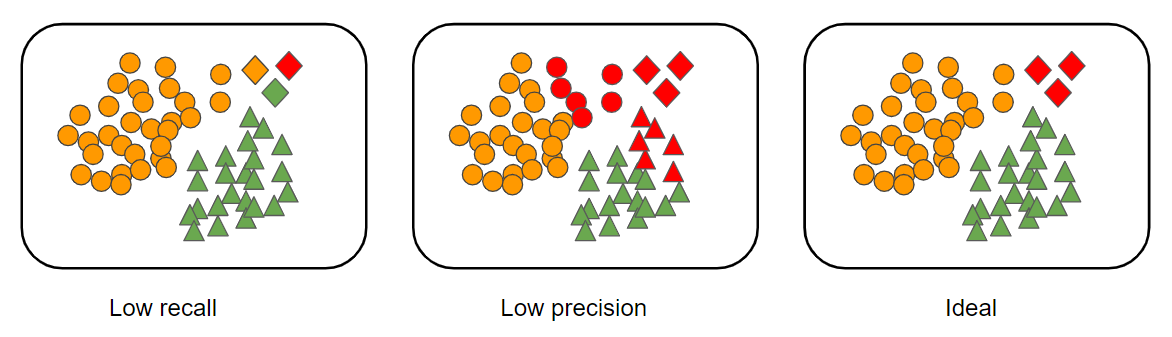}
  \caption{Decision boundary in different situation}
  \label{lowrecall}
\end{figure}

When training from class-imbalanced data, the accuracy of the classifier is mainly hampered by its weak performance for minor classes. There are two sources of this inaccuracy, as schematically illustrated in Figure~\ref{lowrecall}: the low recall --- the situation that samples in minor classes are classified as a major class, or the low precision, where samples from a major class are classified as a minor class. Empirically, we observe when training on the original imbalanced dataset that it is the low recall that affects the prediction accuracy, whereas the precision was relatively high, which is as expected from a small class size. As more unlabelled minor class data are added to the training set, the classifier gradually gains higher recall, but the precision drops as well, which we conjectured is due to the quality of added data, especially those that are added with wrongly predicted labels. This situation is similar to the ``noisy label'' setting described in \cite{MWHZ+2018}, where the training data contains incorrect labels. To improve the recall while keeping the precision from dropping, we also identify samples from the labeled dataset that are wrongly predicted as a minor class, and add them to the training data; these samples are ``hard'' to classify, which suggests they play important roles in characterizing the decision boundary efficiently.




Therefore, in each iteration step of the proposed schema \eqref{obj}, the selected data to be added to training consist of two parts: the unlabeled data from $\mathcal{U}$ and wrongly labeled data from $\mathcal{D}$. We refer to the second part as \emph{context data} as it helps to differentiate minor class samples from their neighboring major class samples. The label of each selected sample from the unlabelled data (in the second part of Formula~\eqref{obj} $\hat{y}_i$) is given by the prediction of $f_c(\textbf{x}_i, \textbf{w}_c)$, while the label for each sample selected from labeled data is used in both the first term and the second term of Formula~\eqref{obj}.


\subsection{SelectNet}
\label{sec:se}

Ideally, the minor class samples in $\mathcal{U}$ should be correctly identified, which will maximize the effect of using extra data. As this rarely occurs in practice, weighing the gain of collecting more information about the minor class samples and the loss of training with noisy labels is critical. This is trade-off is reflected in the second line of Formula~\eqref{obj}, where $f_s$ are indicators of whether a sample should be added into loss calculation or not, and the parameter $\lambda$ balances out the ``gain'' and the conceptual ``loss.'' In previous works such as self-paced training, $f_s$ represents the decision made according to a predefined choosing rule, often resulted from the comparison  between the loss value and a threshold $\lambda$, which is exactly the derivative result of Formula~\ref{spl} with respect to the hyperparameter $\lambda$:

\begin{equation}\label{spcm}
  f_s=\begin{cases}
               1   & \textrm{if }\textbf{L}\left(y_i, f_c\left(\textbf{x}_i, \textbf{w}_c\right)\right) < \lambda\\
               0   & \textrm{otherwise.}
            \end{cases}
\end{equation}
Such a deterministic rule for determining $f_s$ imposes a dependence on the correct choice of hyperparameter $\lambda$, which limits the flexibility of the proposed method. We thus propose a data-driven way, using a regression model to produce $f_s$ for each sample, motivated by the architecture of MentorNet \citep{jiang2017mentornet} which proposed a similar idea for resolving the noisy label issue in supervised learning. In their setting, the extra network is trained on a predefined small set of samples for deciding whether a predicted label is correct or not, which can not be directly used for solving the class imbalanced issue in our problem. The intuition of turning $f_s$ into a data-driven indicator is more on directly expanding the search space of overall target function~\eqref{obj}.

Based on these considerations, we revise the target objective function for training $f_s$ as
\begin{equation}\label{tarfs}
\hat{f}_{si}:=\mathop{\arg \min}_{\Theta} \frac{1}{n_{add}} \sum_{i\in \mathcal{D}'} \hat{f}_s(\textbf{z}_i;\Theta)(\textbf{L}_{ci} - \lambda),
\end{equation}
and $f_s$ is assigned with respect to
\begin{equation}\label{spcm}
  f_{si}=\begin{cases}
               1   &\hat{f}_{si} > \lambda\\
               0   &\textrm{otherwise}
            \end{cases}
\end{equation}
As $f_s$ is trained to minimize the formulation in \eqref{tarfs}, the choice of $\lambda$ is insensitive in a range, for the update of $f_s$ will try to fit to it. In the experiment we train $f_s$ using a simple two-fully-connected-layer network. 

Our final algorithm, the SelectNet, incorporates the ideas of using context data and alternatively updates the selection part $\hat{f}_s$ and the base part of the model on the selected samples. The full algorithm is described in Algorithm~\ref{sel}.

\begin{algorithm}
\caption{SelectNet}

\textbf{Input:} labeled dataset $\mathcal{D}$, unlabeled dataset $\mathcal{U}$, minor classes $\mathcal{C}$\\
\textbf{Output:} classifier with weights \textbf{w}\\
Initialize model \textbf{w} by training on $\mathcal{D}$ with oversampling\\
\textbf{Repeat until convergence:}\\
\textbf{1. }$\mathcal{D}'$=\{prediction($\mathcal{D}\cup\mathcal{U}$, $\textbf{w}^t$) in $\mathcal{C}$\}\\
\textbf{2. }Update $\hat{f}_s^t$ by \{$f_c(\mathcal{D}'$, $\textbf{w}^t$), $L_c(\mathcal{D}'$, $\textbf{w}^t$)\} \\
\textbf{3. }$f_s^t = \textbf{1}_{(prediction(\mathcal{D}', \hat{f}_s^t)>\beta)}$\\
\textbf{4. }$\mathcal{D}_{add}$ = \{$\mathcal{D}'$ when ${f_s}^t$ = 1\}\\
\textbf{5. }$\mathcal{D}_{train}^t$ = $\mathcal{D}\cup\mathcal{D}_{add}$\\
\textbf{6. }Update \textbf{w} using $\mathcal{D}_{train}^t$
\label{sel}
\end{algorithm}

\section{Numerical Experiments}
\label{sec:numer-exper}

We present numerical results to support the proposed SelectNet and compare it with baselines and state-of-the-art algorithms, e.g. the imbalanced training results, methods based on oversampling \citep{chawla2002smote,kubat1997addressing}, the application of self-paced method in imbalanced problem \citep{JMZSH2015,zhou2018sparc}, methods based on context data \citep{zhou2018sparc}, and also a class-balancing loss method\citep{DBLP:journals/corr/abs-1901-05555} as summarized in Section \ref{sec:MC}.

\subsection{Datasets}
CIFAR-10 and CIFAR-100 datasets \citep{krizhevsky2009learning} are adopted throughout this paper. Note that CIFAR-10 and CIFAR-100 contain equal amounts of training samples in each class. Hence, the classification accuracy of the original balanced datasets serves as the upper bound of the achievable classification accuracy.

We artificially create imbalanced datasets using CIFAR-10 and CIFAR-100. In CIFAR-10 experiments, after selecting the minor classes, e.g., classes [0,2,6,7] in our experiments, we create imbalanced datasets with an imbalanced ratio of $90$. In particular, $1\%$ of training samples in the selected minor classes are kept as labeled data and the other $99\%$ of data in the selected minor classes are left as unlabeled data. Similarly, we select classes [10,20,60,70] and [5,10,11,18,30,45,55,79,86,98] as the minor classes. Then, $90\%$ of the data belonging to a major class are kept as labeled data and the other $10\%$ of data in the major class are put into the unlabeled dataset. In CIFAR-100 experiments, we create imbalanced datasets with an imbalanced ratio of $14$ by keeping $5\%$ of training samples in minor classes and $90\%$ of training samples in major classes. 

Recently, a long-tailed CIFAR dataset was proposed in \citep{DBLP:journals/corr/abs-1901-05555} and a new method for imbalanced dataset based on a class-balanced-loss was also proposed in \citep{DBLP:journals/corr/abs-1901-05555}. Hence, we compare SelectNet with the class-balanced-loss method using the long-tailed CIFAR dataset. In this experiment, all the unused training samples are collected as unlabeled samples.


\subsection{Methods for Comparison}
\label{sec:MC}

\paragraph{Imbalanced Training}
In this method, the classifier is only trained with the labeled imbalanced dataset. 

\paragraph{Oversampling}
The oversampling method aims at creating a balanced training dataset using the original imbalanced dataset. Suppose a certain class is minor, it repeatedly samples with replacement from the data in this class and put them together until the number of samples is as large as that of a major class. Then these new samples are added to augment the training data. Such a process is repeated to eliminate minor classes to achieve a balanced training dataset. Finally, a classifier is trained with this new balanced dataset.

\paragraph{Self-paced training}
The idea of self-paced training can be simply migrated to the data imbalance problem. Suppose an unlabeled dataset is available for the self-paced training. After every $n$ epochs of iterations, a certain amount of data is selected from the unlabeled dataset to augment the labeled dataset with labels given by the current classifier. The augmented dataset serves as the new training data for the next $n$ epochs. The selection process is controlled by a threshold $\lambda$, which is set as  $0.6$ in this paper. When an unlabeled sample is predicted as the minor class with a loss smaller than $\lambda$ using the current classifier, this sample will be added to the training dataset.

\paragraph{Context data}
As described in Section \ref{sec:cd}, in addition to the selection of unlabeled data as in the self-paced training, labeled data which are classified as minor classes by the current classifier can also be added to the training dataset for latter training. This approach is named as the context data method. Similar to the self-paced method, a threshold $0.6$ is set to control the selection process. When a labeled sample is predicted as the minor class with a loss smaller than $\lambda$, this sample will be duplicated and added to the training dataset.

\paragraph{SelectNet}
As described in Section \ref{sec:se}, after every $n$ epochs, a deep neural network is trained to decide whether a sample should be added in the next stage of training. The model we use is a fully-connected ReLU neural network with two hidden layers of width $8$ and $4$, respectively. The output of this network is a $1$-d scalar from a Sigmoid activation. If the output of the network is larger than a hyperparameter $\lambda=0.6$, the corresponding input sample will be added to the training dataset for later iterations.

For the first two methods, we set the number of epochs to be $200$, while for the last three methods, we update the training samples every $10$ epochs for $20$ iterations, i.e., the total number of epochs is also $200$. 

\subsection{Comparison on CIFAR imbalanced dataset}

In this experiment, we show the universal advantage of the proposed SelectNet over various deep learning methods like the imbalanced train method, the oversampling method, the self-paced method, and the context data method. To make consistent comparisons, we fix the same kind of deep neural network to implement different classification methods above. To see the influence of different network architectures on the performance of the classification methods, commonly used architectures have been explored in the test, e.g., the standard network example in Keras \citep{chollet2015keras}, ResNet-20, and ResNet-56.

Table \ref{big-acc} summarizes the experiment results for CIFAR imbalanced datasets. Numerical results show that SelectNet is almost consistently better than other methods for both CIFAR-10 and CIFAR-100 datasets and various network structures. There is only one case in which SelectNet is the second best method with accuracy only slightly smaller than that of the self-paced method when the network is ResNet20 and the dataset is CIFAR-100. 

Note that SelectNet is built on top of the context data method and an extra deep neural network for adaptively update training data. Hence, the results of the SelectNet is always better than the context data method, the results of which is already very promising. The consistent advantage of the SelectNet over the context data method validates the proposal of an extra deep neural network for updating training data adaptively.

Table~\ref{catperf} shows the category-wise f1-score of three methods for CIFAR-10 in the simple net experiment. The bold numbers are the performance of the minor classes. Each of these suffers poor accuracy when using the oversampling method. The latter two methods, with the help of additional data, considerably increased the performance of minor classes. Meanwhile, our propose SelectNet exhibits the best improvements.

\begin{table}
  \caption{CIFAR accuracy comparison}
  \label{big-acc}
  \centering
  \begin{tabular}{lllllll}
    \toprule
    &\multicolumn{3}{c}{CIFAR-10}   &   \multicolumn{3}{c}{CIFAR-100}     \\
    \midrule
    &\multicolumn{3}{c}{4 minors}&\multicolumn{2}{c}{4 minors}  & \multicolumn{1}{c}{10 minors}\\
    Method     & Simple net & ResNet20 &  ResNet56 & ResNet20 &  ResNet56 &  ResNet56\\
    \midrule
    Imbalanced train & 0.5617  & 0.5659  & 0.5672 &0.5729   & 0.5806  &  0.589    \\
    Oversampling     & 0.5744 &  0.6629 & 0.6897  & 0.5749 &  0.6128 &   0.6014       \\
    Self-paced     & 0.6933       & 0.7894  & 0.7951 & \textbf{0.5971}   & 0.5901  & 0.5974  \\
    Context data & 0.7403 & 0.7881 & 0.7864 & 0.5889 & 0.6285 & 0.6165\\
    SelectNet & \textbf{0.7424} & \textbf{0.7895} & \textbf{0.8011} & 0.5921 & \textbf{0.6290} & \textbf{0.6171}  \\

    \bottomrule
  \end{tabular}
\end{table}

Table \ref{big-acc} compares various methods in terms of the overall classification accuracy for all classes, while Table \ref{catperf} compares these methods in terms of the classification accuracy within individual classes. It is worth emphasizing that the SelectNet significantly outperforms other methods in the classification of minor classes, which is of special interest in real applications, especially in medical applications where minor classes are more valuable.

\begin{table}
  \caption{CIFAR-10: class-wise prediction accuracy.}
  \label{catperf}
  \centering
  \begin{tabular}{lllllllllll}
    \toprule
&\textbf{0}&1&\textbf{2}&3&4&5&\textbf{6}&\textbf{7}&8&9\\
    \midrule
Oversampling&\textbf{0.05}&0.85&\textbf{0.02}&0.41&0.49&0.50&\textbf{0.06}&\textbf{0.09}&0.68&0.79\\
Self-paced&\textbf{0.71}&0.90&\textbf{0.20}&0.52&0.71&0.64&\textbf{0.52}&\textbf{0.72}&0.86&0.86\\
SelectNet&\textbf{0.74}&0.89&\textbf{0.54}&0.57&0.71&0.69&\textbf{0.72}&\textbf{0.76}&0.86&0.85\\
    \bottomrule
  \end{tabular}
\end{table}

\begin{figure}
  \centering
  \includegraphics[width=1\linewidth]{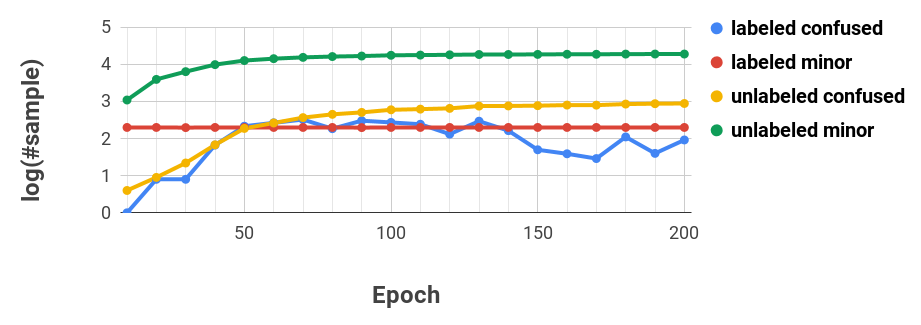}
  \caption{Number of data chosen during training.}
  \label{number}
\end{figure}

To check the number of extra training samples added during the training of SelectNet, we choose the case when networks are carried out via the simple one in Keras \citep{chollet2015keras} and visualize the numbers in Figure~\ref{number}. The ``labeled-confused" number represents the number of labeled samples that are wrongly predicted  (no matter major or minor) and added to the training dataset.  The ``labeled-minor" number denotes the number of labeled samples that are correctly classified as a minor sample. Similarly, the ``unlabeled-confused" and ``unlabeled-minor" numbers mean the numbers in the case of unlabeled samples. These numbers seem to be bounded by a constant number, which means that the mistakes SelectNet makes would not increase after a certain number of epochs. The ``unlabeled-minor" number is significantly larger than other numbers, indicating the effectiveness of the SelectNet since most of the selections it makes are correct.

\subsection{Comparison on long-tailed CIFAR dataset}
Here, we compare SelectNet with the newly proposed class-balanced-loss method in  \citep{DBLP:journals/corr/abs-1901-05555} using the long-tailed CIFAR dataset therein. Since there is no class that has a number of samples significantly smaller than others, we treat half classes with a smaller amount of data as the minor classes. In particular, in the long-tailed CIFAR-10 dataset, the minor classes are 5,6,7,8,9; while in the case of long-tailed CIFAR-100, the minor classes are 50 to 99. All the other settings remain the same as in previous experiments. The comparison is conducted with ResNet32 and the results are summarized in Table \ref{longtail}. Obviously, the proposed SelectNet outperforms the class-balanced-loss method by $1\%$ classification accuracy.

\begin{table}
  \caption{Long-Tailed CIFAR accuracy comparison.}
  \label{longtail}
  \centering
  \begin{tabular}{llll}
    \toprule
    Method     & CIFAR-10-0.01 &  CIFAR-100-0.01\\
    \midrule
    Class-Balanced Loss & 0.7457 & 0.3960\\
    SelectNet & \textbf{0.7576} & \textbf{0.4056}\\
    \bottomrule
  \end{tabular}
\end{table}

\section{Conclusion}
\label{sec:concl-future-work}

In this work, we drew an analogy between the imbalanced data problem and semi-supervised learning, and proposed a simple yet powerful approach, referred to as SelectNet, to mitigate the class imbalance issue using unlabeled data ``in the wild.'' We began with the observation that incorporating ``context data'' into training significantly improves the classification performance, then generalized the $0$-$1$ selection rule to a continuously valued regression network that takes real values between $0$ and $1$ as ``selection.'' Combining the idea of context data and minor class data selection provides significant improvement of the classification performance over existing works. We expect other cost-sensitive learning techniques to benefit from this ``data side improvement'' as well; we will explore this in future work. The code of this paper will be available in the authors' personal homepage.






\newpage

\small
\bibliographystyle{abbrvnat}
\bibliography{refs}   




\end{document}